\documentclass[letterpaper]{article} 
\usepackage{aaai23}  
\usepackage{times}  
\usepackage{helvet}  
\usepackage{courier}  
\usepackage[hyphens]{url}  
\usepackage{graphicx} 
\usepackage{natbib}  
\usepackage{caption} 
\usepackage{microtype}
\usepackage{graphicx}
\usepackage{caption}
\usepackage{subcaption}
\usepackage{booktabs} 
\usepackage{multirow}
\usepackage{amsmath,amsfonts}
\usepackage{algorithmic}
\usepackage{algorithm}
\usepackage{array}

\usepackage{textcomp}
\usepackage{stfloats}
\usepackage{url}
\usepackage{verbatim}

\usepackage{cite}

\usepackage[utf8]{inputenc}

\usepackage{longtable}
\usepackage{tabularx}

\usepackage{ulem}

\usepackage{amsmath}
\usepackage{stmaryrd}

\usepackage{amssymb}
\usepackage{verbatim}

\usepackage{color}
\usepackage{bm}
\usepackage{ulem}
\usepackage{newfloat}
\usepackage{listings}
\DeclareCaptionStyle{ruled}{labelfont=normalfont,labelsep=colon,strut=off} 
\floatstyle{ruled}
\newfloat{listing}{tb}{lst}{}
\floatname{listing}{Listing}
\title{FedABC: Targeting Fair Competition in Personalized Federated Learning}
\author {
    Dui Wang\textsuperscript{\rm 1,\rm 2,\rm 3}\footnote{This work was done when Dui Wang was a research intern at JD Explore Academy.},
    Li Shen \textsuperscript{\rm 3},
    Yong Luo\textsuperscript{\rm 1,\rm 2},
    Han Hu \textsuperscript{\rm 4},
    Kehua Su\textsuperscript{\rm 1}\footnote{Corrsponding Author.},
    Yonggang Wen\textsuperscript{\rm 5},
    Dacheng Tao \textsuperscript{\rm 3}
}
\affiliations {
    \textsuperscript{\rm 1} National Engineering Research Center for Multimedia Software, School of Computer Science, Institute of Artificial Intelligence and Hubei Key Laboratory of Multimedia and Network Communication Engineering, Wuhan University, China,\\
    \textsuperscript{\rm 2} Hubei Luojia Laboratory, Wuhan, China,
    \ \textsuperscript{\rm 3} JD Explore Academy, China,\\
    \textsuperscript{\rm 4} School of Information and Electronics, Beijing Institute of Technology, China,\\
    \textsuperscript{\rm 5} School of Computer Science and Engineering, Nanyang Technological University, Singapore,\\
    \{wangdui, luoyong, skh\}@whu.edu.cn,\\
    mathshenli@gmail.com,
     hhu@bit.edu.cn,
     ygwen@ntu.edu.sg,
     dacheng.tao@gmail.com
}

\begin{document}

\maketitle

\begin{abstract}
Federated learning aims to collaboratively train models without accessing their client's local private data. The data may be Non-IID for different clients and thus resulting in poor performance. Recently, personalized federated learning (PFL) has achieved great success in handling Non-IID data by enforcing regularization in local optimization or improving the model aggregation scheme on the server. However, most of the PFL approaches do not take into account the unfair competition issue caused by the imbalanced data distribution and lack of positive samples for some classes in each client. To address this issue, we propose a novel and generic PFL framework termed Federated Averaging via Binary Classification, dubbed FedABC. In particular, we adopt the ``one-vs-all'' training strategy in each client to alleviate the unfair competition between classes by constructing a personalized binary classification problem for each class. This may aggravate the class imbalance challenge and thus a novel personalized binary classification loss that incorporates both the under-sampling and hard sample mining strategies is designed. Extensive experiments are conducted on two popular datasets under different settings, and the results demonstrate that our FedABC can significantly outperform the existing counterparts.
\end{abstract}

\section{Introduction}
Federated learning (FL) is an emerging machine learning  paradigm that trains an algorithm across multiple decentralized clients (such as edge devices) or servers without exchanging local data samples~\cite{mcmahan2017communication}. In this big data era, large-scale data are becoming increasingly popular, but also suffer the risk of data leakage. FL aims at addressing this issue by letting the clients update models using private data and the server periodically aggregate these models for multiple communication rounds. Such decentralized learning has shown its great potential to facilitate real-world applications, including healthcare~\cite{xu2021federated}, user verification~\cite{hosseini2021federated} and the Internet of Things (IoT)~\cite{zheng2022applications,huang2022stochastic}.

A key challenge in federated learning is the training given non-independent and identically distributed (non-i.i.d.) data~\cite{hsieh2020non}. Due to the varied data distributions among different clients, the single global model (GM) obtained by simple averaging is hard to cater for all heterogeneous clients. Besides, clients update the global model on their local dataset, making local model misaligned and leading to weight divergence~\cite{zhao2018federated}. These issues have been found to result in unstable and slow convergence~\cite{li2020federated} and even extremely poor performance~\cite{li2021federated}.

\begin{figure}[!t]
    \begin{center}
    \includegraphics[width=1\columnwidth]{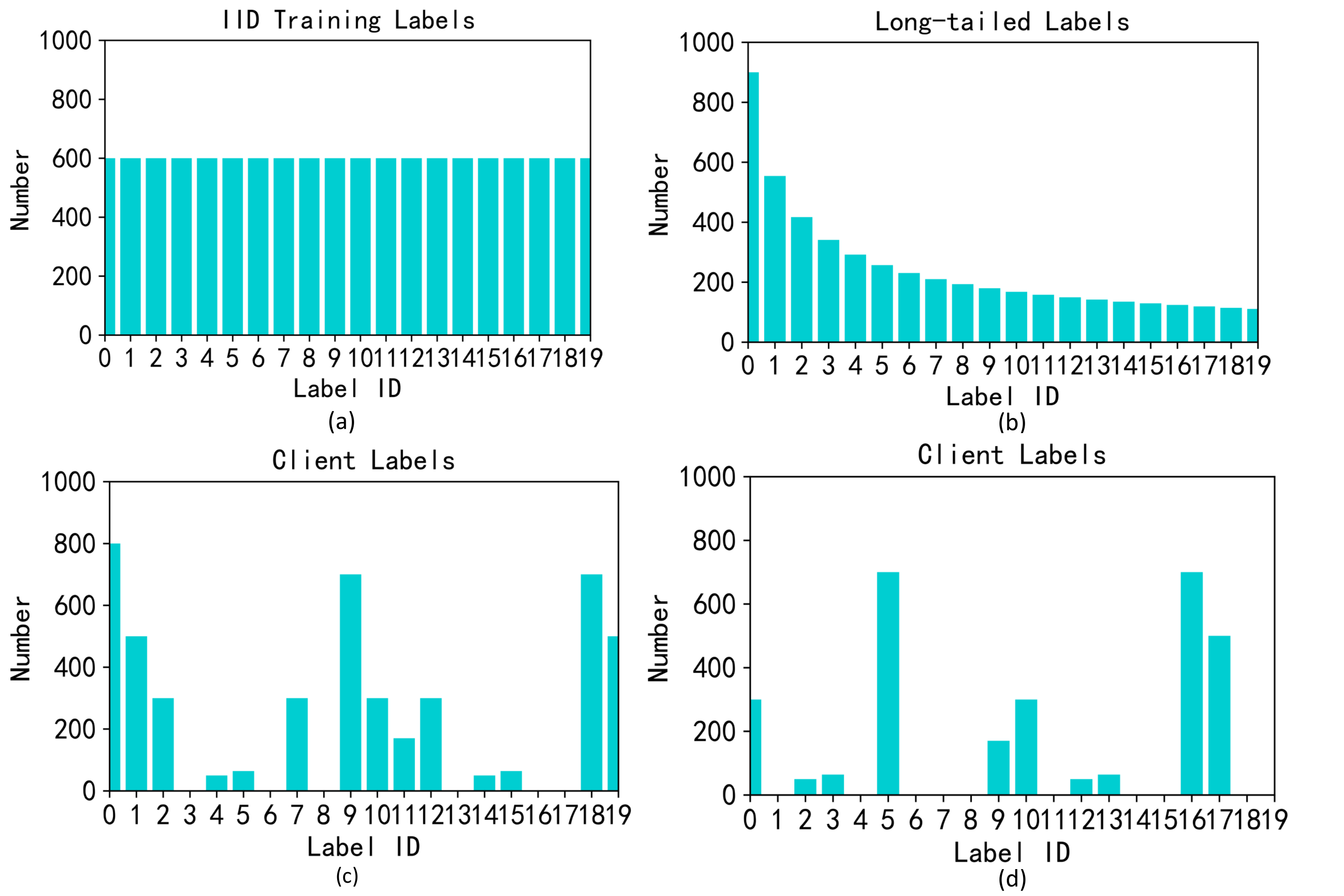}
    \end{center}
    \caption{An illustration of different situations of data distributions in FL: (a) ideal i.i.d; (b) long-tail; (c) and (d) lack of positive samples for some classes.}
    \label{fig:1}
\end{figure}

To deal with heterogeneity in FL, numerous solutions have been proposed, such as the ones that constrain the direction of local model update to align the local and global optimization objectives~\cite{li2020federated,karimireddy2020scaffold,acar2021federated,zhang2022fine,liu2022deep}. Personalized federated learning (PFL)~\cite{smith2017federated,huang2022achieving,dai2022dispfl} is a promising solution to addresses this challenge by jointly learning multiple personalized models (PMs), one for each client. For instance, references~\cite{collins2021exploiting,liang2020think,sun2021partialfed} exploit flexible parameter sharing strategies that only transmit partial model parameters, and the local models are regularized using a learnable global model~\cite{t2020personalized,hanzely2020lower,li2021ditto}, \citet{dai2022dispfl,huang2022achieving} adopt sparse training to achieve personalization.

\begin{figure}[!t]
    \begin{center}
    \includegraphics[width=1\columnwidth]{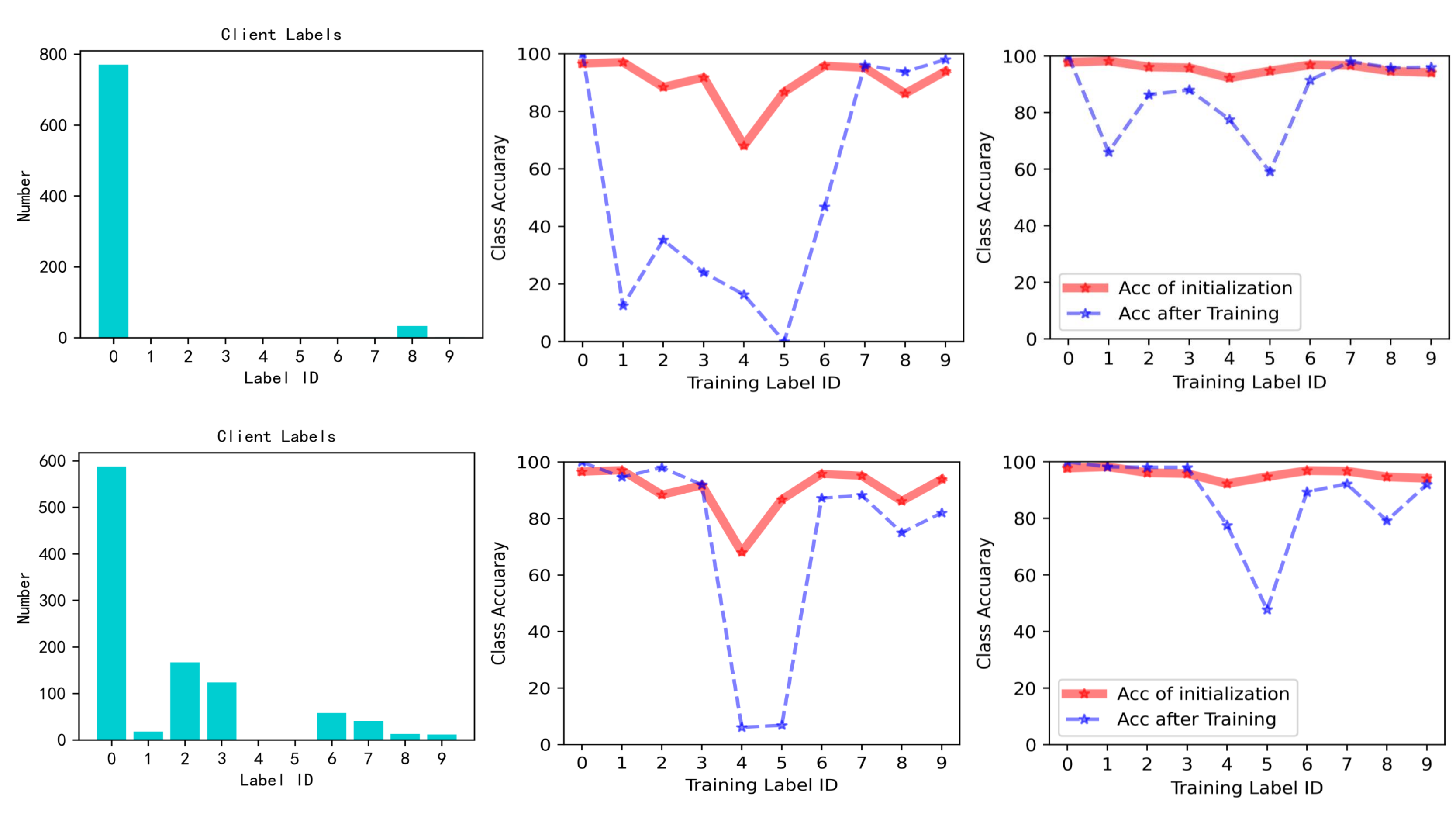}
    \end{center}
    \caption{An illustration of the poor performance for some classes that have few or no positive samples. The middle and last columns are results after $10$ and $50$ rounds of global aggregation and local updating, respectively.}
    \label{fig:2}
\end{figure}

However, all these approaches merely conduct an extra regularization or improve the aggregation strategy, which ignore some extraordinary situations of data distribution (see Figure~\ref{fig:1}) that often exist in FL clients. The situations can be divided into two categories: 1) imbalanced class distribution~\cite{he2009learning}, where some classes have much more/less samples than others; 2) lacking positive samples for some classes,
and the probability of occurrence increases with the degree of heterogeneity.
In these extreme cases, if the ubiquitous Softmax function (together with cross-entropy loss)~\cite{jang2016categorical} is adopted, the normalization would enforce all the class logits summed to one, and thus induce competition among different classes. That is, increasing the confidence of one class will necessarily decrease the confidence of some other classes.
This can easily lead to over-confident predictions~\cite{guo2017calibration} for dominated classes, sub-optimal performance for minority classes and extremely poor performance for classes that lack positive samples (see Figure~\ref{fig:2}). Although communicating with the server can alleviate this issue to some extent, the neural weights of each model can be randomly permuted and thus hard to fully assemble knowledge of all clients. Besides, the classes that have poor performance in local models are hard to achieve promising global performance by indiscriminately point-wise-averaging.

To address the unfair competition between classes in FL clients,
we propose a novel method that boosts the performance of standard PFL termed \textbf{Fed}erated \textbf{A}veraging via \textbf{B}inary \textbf{C}lassification (FedABC), which adopts the well-known ``one-vs-all'' strategy~\cite{rifkin2004defense,wen2021sphereface2} to reduce the unfair competition among classes and focus more on personalized classes. Different from the traditional multi-class classification training based on Softmax, our FedABC performs binary classification for each category, where the feature extractors are shared for different classes.
In particular, given $K$ classes in the training set, FedABC constructs $K$ binary classification problems, where data from the target class are treated as positive samples and data from the remaining classes are treated as negative samples. By employing this strategy, the classes that have few or even no positive samples will not be suppressed by the majority categories in the prediction, and thus can be liberated from unfair competition.
This enables us to focus on each class and personally deal with its issue of either data imbalancing or lacking of positive samples,
which is tackled by designing a novel and effective binary loss function that incorporates both the under-sampling~\cite{yen2009cluster} and hard-sample mining~\cite{schroff2015facenet,wu2017sampling} strategies. These strategies focus on the learning of hard samples and reducing the impact of easy samples. 

To summarize, the main contributions of this paper are:
\begin{itemize}
  \item We propose a novel FL method termed FedABC that adopts a binary classification strategy to increase personality of the learning for each class and liberate the different classes from unfair competition for the heterogeneous clients;
  \item We design an effective binary loss function to alleviate the imbalanced data and insufficient positive sample issues by incorporating both the hard-sample mining and under-sampling strategies.
\end{itemize}
We conduct extensive experiments on two popular visual datasets (CIFAR-10 and MNIST) under four heterogeneity settings. The results demonstrate the effectiveness of the proposed FedABC over the competitive baselines.


\section{Related work}

\paragraph{Federated learning} is a machine learning paradigm that aims to collaboratively learn a model via
coordinated communication with multiple clients, which do not access to the client's local data. FedAvg~\cite{mcmahan2017communication}, a well-known FL algorithm, learns a global model by simple averaging the local models. A variety of recent works have proposed for FL and achieved promising results. For example, ~\cite{li2020federated,karimireddy2020scaffold,acar2021federated} add regularization term to restrain the client's training process from moving too far and thus alleviate client drift. Personalized Federated Learning (PFL)~\cite{smith2017federated} addresses this challenge by jointly learning multiple personalized models, one for each client. For example, ~\cite{collins2021exploiting,liang2020think,sun2021partialfed} exploit flexible parameter sharing strategies that only transmitting partial model parameters between clients and server. Besides, local fine-tuning~\cite{wang2019federated},meta-learning~\cite{chen2018federated,khodak2019adaptive} and multi-task learning~\cite{smith2017federated} are also introduced in PFL.

\begin{figure*}[t]
    \begin{center}
    \includegraphics[width=2\columnwidth, height=0.5\textwidth]{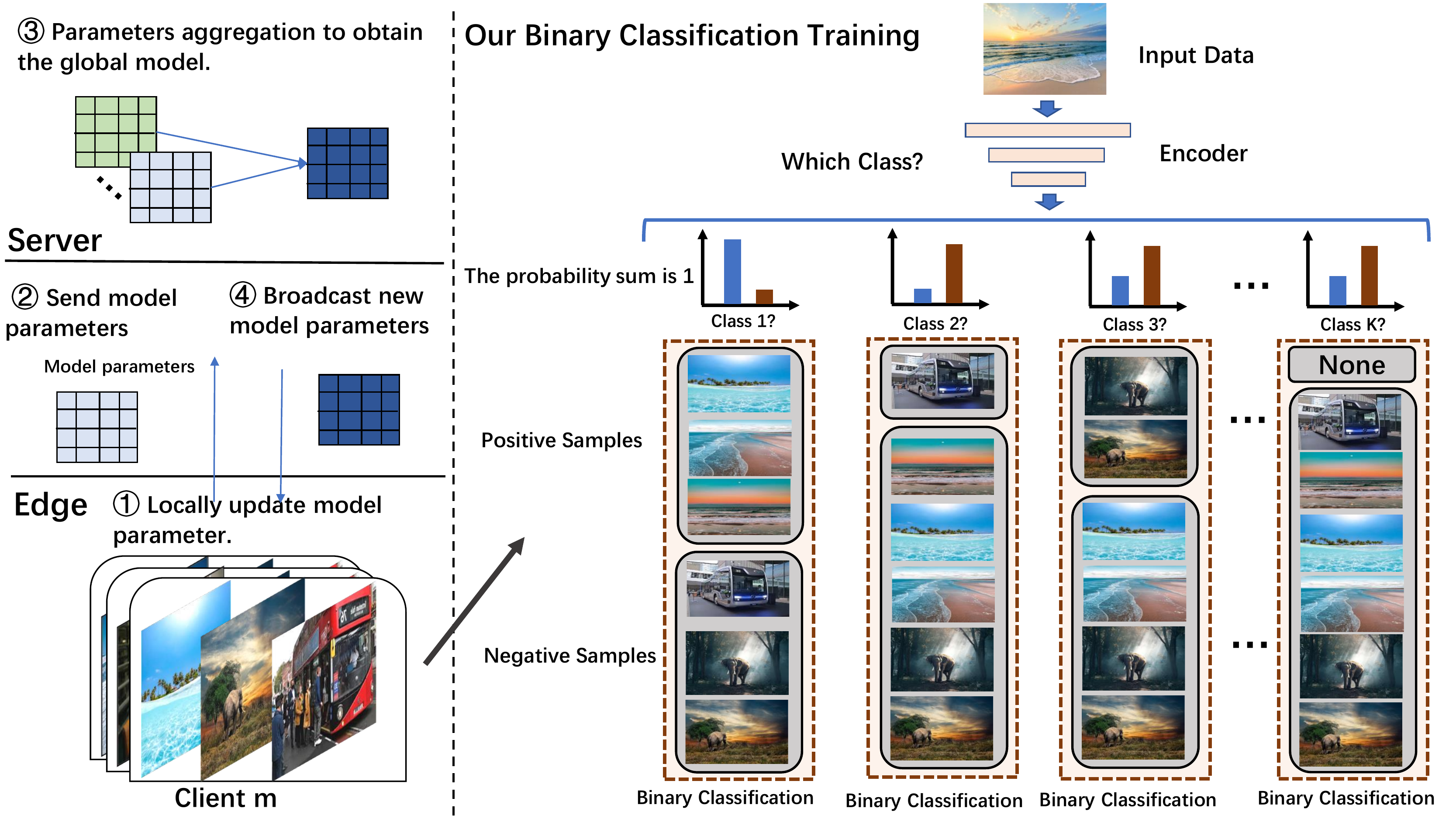}
    \end{center}
    \caption{\textbf{Overview of the proposed federated averaging via binary classification (FedABC) framework}, which is based on the vanilla federated learning paradigm, where the server orchestrates the learning amongst clients and is responsible for aggregating the client personalized model parameters. In the local training process, FedABC adopts the ``one-vs-all'' training strategy that learns an independent binary classifier for each category, while the feature extractor is shared. This may lead to severe class imbalancing problem, and some classes may have no positive samples, and therefore a novel personalized binary classification loss is designed.}
    \label{pipeline}
\end{figure*}

\paragraph{Imbalanced classification}~\cite{huang2016learning} aims to train a model on the dataset where a few class occupy most instances and remaining classes have few instances. In such case, typical models perform biasedly toward the majority classes and poorly on the minority classes~\cite{japkowicz2002class,he2009learning,van2017devil,buda2018systematic}. To address this challenge, a numerous works have been proposed and can be divided into two categories in general: 1) re-sampling~\cite{shen2016relay,geifman2017deep,zou2018unsupervised}, which usually employs over-sampling~\cite{kang2019decoupling} for the minority class or under-sample~\cite{drummond2003c4} for the majority class to re-balance data distribution. Over-sampling adds repeated samples for the minority class and sometimes causes over-fitting. A recently work ~\cite{katharopoulos2018not} shows that most samples of the majority class contribute less for later model training, such as the easy-samples; 2) cost-sensitive re-weighting ~\cite{aurelio2019learning,hong2021disentangling,ren2020balanced}, which assigns important-weight for samples to increase the occupy of minority class and reduce the occupy of majority class. For example, some methods assign weight by using the inverse square root of class frequency or its smooth version. Besides, the Focal loss~\cite{lin2017focal} is a classic solution for the imbalanced classification problem, and the main idea is to focus on learning the hard sample and reduce the impact of easy sample.

\section{Federated Averaging via Binary Classification}

In this section, we present the proposed FedABC, which is based on the vanilla federated by averaging framework, where a central server exchange averaging model parameter with clients as
depicted in the left part of Figure~\ref{pipeline}. In the client training process, different from the traditional training approaches that directly adopt the softmax function for multi-class classification, our FedABC adopts the ``one-vs-all'' training strategy as illustrated in the right part of Figure~\ref{pipeline}. This not only alleviates the unfair competition between classes, but also enables us to focus on each category to design personalized loss function for each client. This facilitates the tackling of the severe class imbalance issue. More details are depicted as follows.

\subsection{Problem formulation}
In this work, we consider a typical PFL setting for supervised learning, i.e., the multi-class classification. We suppose that there are $m$ clients and $C$ classes in total. For $i \in [m]$, the $i$-th client has individual data distribution $\mathcal{D}_{i}$, where some classes have many samples and the remaining classes have much fewer. The classes that have positive samples are denoted as $C_{i}^{p}$, and some classes that have no positive samples are denoted as $C_{i}^{n}$. %
The ${i}$-th client has access to the local dataset $\mathcal{S}^{i}=\{({x}_{i}^{1},y_{i}^{1}),({x}_{i}^{2},y_{i}^{2}),\cdots,({x}^{n_{i}}_{i},y_{i}^{n_{i}})\} $. %
The neural network parameter $\boldsymbol{\theta}_{i}$ of the $i$-th client consists of two parts: an embedding network $f: \mathcal{X} \rightarrow \mathcal{Z}$ parameterized by $\boldsymbol{\theta}^{f}_{i} $ maps the input ${x}$ to the latent feature, i.e., ${z}:= f({x};\boldsymbol{\theta}_{i}^{f})$, and a predictor $h:\mathcal{Z} \rightarrow \mathcal{Y}$ parameterized by $\boldsymbol{\theta}^{h}_{i}$ maps latent feature to the logits ${y}$ , i.e.,  $y:=h(f({x}; \boldsymbol{\theta}_{i}^{f}); \boldsymbol{\theta}_{i}^{h})$.
The corresponding feature extractor part and predictor part of the global model $\theta$ achieved by averaging are defined as $\boldsymbol{\theta}^{f}$ and $\boldsymbol{\theta}^{h}$, respectively. A non-linear activation $\sigma(\cdot)$ is applied to $y$, so the output of the neural network can be defined as $\sigma(h(f({x}; \boldsymbol{\theta}^{f}), \boldsymbol{\theta}^{h}))$, and we rewrite it as $q:=\sigma({g}({x};\boldsymbol{\theta}^{f};\boldsymbol{\theta}^{h}))$. %
Since the binary classification is employed, we adopt the $\textbf{sigmoid}$ function for activation, which maps the input into $[0,1]$. For $i\in[m],j\in[C]$, the binary loss for the corresponding class can be defined as $\ell_{i}^{j}(x_{i},y_{i};\boldsymbol{\theta})$. We formulate PFL problem according to \cite{hanzely2021personalized} into the following optimization problem:
\begin{equation}
\begin{gathered}
\small
\min _{\left\{\boldsymbol{\theta}_{1}, \cdots, \boldsymbol{\theta}_{m}\right\}} f\left(\boldsymbol{\theta}_{1}, \cdots, \boldsymbol{\theta}_{m}\right)=\frac{1}{m} \sum_{i=1}^{m}F_{i}\left(\boldsymbol{\theta}_{i}\right), \\
F_{i}\left(\boldsymbol{\theta}_{i}\right):=\mathbb{E}\left[\mathcal{L}_{({x}, y) \sim \mathcal{D}_{i}}\left(\mathcal{S}^{i};\boldsymbol{\theta}_{i} \right)\right].
\end{gathered}
\label{general-PFL}
\end{equation}
PFL learns a model $\boldsymbol{\theta_{i}}$ for the $i$-th client, the goal of which is to perform well on the local dataset $\mathcal{S}_{i}$, and the local model parameters $\boldsymbol{\theta_{i}}$ are often initialized with the global model $\boldsymbol{\theta}$. In PFL, many existing works add a regularizer either on the server-side to improve the aggregation scheme or on the client-side to improve the local optimization, but there is no agreed objective function so far. Most of the existing generic FL works can be easily adopted in PFL without extra processing. This is achieved by utilizing the locally trained model as its personalized model.
In our work, we propose a novel method that adopts binary training strategy to tackle the imbalanced problem and boost the generalization of client model. We do not need any extra information or other proxy data, and the objective can be formulated by Eq.~(\ref{general-PFL}).

\subsection{FL binary loss function}
The goal of our method is to train a efficient binary classifier for each class, a naive binary loss formulation of class $c$ from the cross entropy (CE) loss is given by following:
\begin{equation}
    L_{\textit{BCE}}(c,q,y)=-[y{\log}(q)+(1-y){\log}(1-q)],
\label{BCE}
\end{equation}
where probability $q:=\sigma({g}(x;\boldsymbol{\theta}^{f};\boldsymbol{\theta}^{h})$, and $q\in[0,1]$ is the output after the operation of sigmoid activation, and $y = 0, 1$ is the samples true label. According to the class situation mentioned above, we rewrite the binary loss function (\ref{BCE}) into the following formulation:
\begin{equation}
L_{\textit{BCE}}(c,q,y)=\left\{
\begin{aligned}
&-[y{\log}(q)+(1-y){\log}(1-q)], & c \in C^{p} \\
&-{\log}(1-q). & c \in C^{n}\\
\end{aligned}
\right.
\label{FL_BCE}
\end{equation}
Here, we neglect the positive part loss item for $C^{n}$ due to the lack of positive samples. By employing this binary training framework, classes can avoid the enforced and competitive normalization induced by the ``softmax'' operation, and thus the unfair competition can be liberated to a certain extent. However, the imbalance problem within each binary classifier may become more serve due to the large amounts of negative samples and few or even no positive samples for the corresponding category in most cases. To address this problem, we propose to incorporate the under-sampling and hard sample mining strategies into the loss function.

\paragraph{Incorporating under-sampling.}
The under-sampling strategy aimed to abandon low-value samples and re-balance data distribution to alleviate the imbalancing issue. In particular, we quantity the significance of different samples according to the output probability $q$. That is, the model will add samples, whose importance values are larger than the pre-defined threshold, into the current training batch, and the remained ones will be abandoned directly. This process is dynamic for each training epoch. The formulation of binary loss function that incorporates the under-sampling strategy is defined as:
\begin{equation}
\begin{split}
L_{\textit{BCE}}^{p}(q,y)&=\left\{
\begin{aligned}
&-{\log}(q), & y=1, q<m^{p} \\
&-{\log}(1-q), & y=0, q>m^{n}\\
\end{aligned}
\right.\\
L_{\textit{BCE}}^{n}(q,y)&=-{\log}(1-q). \quad \quad  y=0, q>m^{nn}
\end{split}
\label{BCE_under_sampling}
\end{equation}

The binary loss function for $C^{p}$ is denoted as $L_{\textit{BCE}}^{p}(c,p,y)$, while $L_{\textit{BCE}}^{n}(c,p,y)$ signifies the loss for $C^{n}$, which only has negative samples. Some positive samples may have already been predicted correctly with probability near to one, and thus contribute little in the future training. The positive samples with low probability are more valuable and need to be better trained in this epoch. Conversely, we maintain the negative samples whose $q$ is larger than a certain threshold. Three hyper-parameters $m^{p}$, $m^{n}$, and $m^{nn}$ should be determined for each class, and the same hyper-parameters are adopted for different clients for simplicity.

\paragraph{Incorporating hard sample mining.}
To further alleviate the imbalancing issue, we incorporate the strategy for mining hard samples into the loss. The well-known Focal Loss~\cite{lin2017focal} is widely-used loss that utilizes the hard sample mining strategy to re-balance the loss contribution of easy samples and hard samples. We follow this hard sample mining strategy to alleviate the imbalancing problem by focusing more on hard samples and lowering the significance of easy samples. This leads to the following final formulation for our final binary loss function:
\begin{equation}
\begin{split}
L_{\textit{BCE}}^{p}(q,y)&=\left\{
\begin{aligned}
&-(1-q)^{\sigma}{\log}(q), & y=1, q<m^{p} \\
&-q^{\sigma}{\log}(1-q), & y=0, q>m^{n}\\
\end{aligned}
\right.\\
L_{\textit{BCE}}^{n}(q,y)&=-q^{\sigma}{\log}(1-q), \quad \quad \quad y=0, q>m^{nn}\!\!\!
\end{split}
\label{BCE_final}
\end{equation}
where $\sigma$ is a hyper-parameter to control the degree of hard sample mining.

\begin{small}
\begin{algorithm}[!t]
	\caption{Federated Averaging via Binary classification}
	\label{alg}
	\begin{algorithmic}
		\STATE \textbf{Input:} $m$ clients, $\mathcal{S}^{i}$ at the $i$-th client.
		\STATE Server initializes parameters $\boldsymbol{\theta}^{0}$.
		\STATE Server sends the initialization to clients.
	    \STATE \textbf{for} t=0,1,2,...,T-1 \textbf{do}:
	    \STATE \quad Server sends $\theta^{t}$ to the $i$-th client
	    \STATE \quad \textbf{for} i=1,2,...,C \textbf{do}
	    \STATE \quad \quad $\theta^{t}_{i} \leftarrow\boldsymbol{\theta^{t}}_{i}, -\eta \nabla_{\boldsymbol{\theta^{t}}} \hat{\mathcal{R}}\left(\mathcal{S}^{i},\boldsymbol{\theta}_{i}\right)$
	    \STATE \quad \quad $\text{where } \hat{\mathcal{R}}\left(\mathcal{S}^{i},\boldsymbol{\theta}_{i}\right)=\mathbb{E}\left[\mathcal{L}^{\textit{BCE}}_{({x}, y) \sim \mathcal{D}_{i}}\left(\mathcal{S}^{i};\boldsymbol{\theta}_{i} \right)\right]$

	    \STATE \quad \quad
	    The $i$-th client maintain $\boldsymbol{\theta}_{i}^{t}$ as the current PM.
	    \STATE \quad \quad
	    The $i$-th client sends $\boldsymbol{\theta}^{t}_{i}$ to the server.
	    \STATE \quad \textbf{end for}
	    \STATE \quad Server updates the model parameters by averaging:\\
	    \quad $\boldsymbol{\theta} ^{t+1}=\sum\limits_{i\in [C] } \frac{|\mathcal{S}^{i}|}{|\mathcal{S}|} \boldsymbol{\theta}^{t}_{i}$
	    \STATE \textbf{end for}
	    \STATE \textbf{Output:} PM$:\{\sum_{i}^{m}\boldsymbol{\theta_{i}}^{\mathrm{T}}\}$, GM$:\{\boldsymbol{\theta}^{\mathrm{T}}\}$
	\end{algorithmic}
\end{algorithm}
\end{small}

\subsection{Training strategy}

Our method adopts the binary training strategy in the local learning process. Firstly, an encoder including an embedding network and a classifier maps the input data into the low-dimensional representation. Then the \textbf{sigmoid} activation is applied to get the final logit, which lies in the interval $[0,1]$. Our method trains an independent classifier for each category to complete the binary classification task and the feature extractor is shared, more details are depicted in Figure~\ref{pipeline}. In practice, the classifier can be any neural network, and the output is just a scalar logit for the corresponding class. Our method does not need any extra modifications on the structure of the neural network or other auxiliary information, and we only need to apply sigmoid activation and adopt our proposed binary loss function presented in Eq.~(\ref{BCE_final}).
The empirical loss of the $i$-th client on the local dataset $\mathcal{S}^{i}$ can be given by:
\begin{equation}
\begin{split}
\mathbb{E}\left[\mathcal{L}^{\textit{BCE}}_{({x}, y) \sim \mathcal{D}_{i}}\left(\mathcal{S}^{i};\boldsymbol{\theta_{i}} \right)\right]:=&\frac{1}{|\mathcal{S}^{i}|} \left(\sum_{j}^{C_{i}^{p}}{L}^{p}_\textit{BCE}(\mathcal{S}^{i};\boldsymbol{\theta}_{i})\right.\\
&\left.+\sum_{j}^{C_{i}^{n}} L_{\textit{BCE}}^{n}(\mathcal{S}^{i};\boldsymbol{\theta}_{i}) \right).
\end{split}
\end{equation}
Since our binary loss function have different forms w.r.t. $C_{i}^{p}$ and $C_{i}^{p}$, we do not merge them together. Employing this training strategy can liberate unfair competition between classes in the classifier and be more focused on personalized classes.
Although there may exist severe problems of data imbalancing and lacking positive samples, we can adopt the customized binary loss function Eq.~(\ref{BCE_final}) to significantly alleviate it. In Algorithm~\ref{alg}, we summarize the learning procedure of our proposed FedABC method.

\section{Experiments}

\subsection{Setup}

\begin{figure*}[!t]
    \begin{center}

        \begin{subfigure}[b]{0.23\textwidth}
            \centerline{\includegraphics[width=1.2\columnwidth]{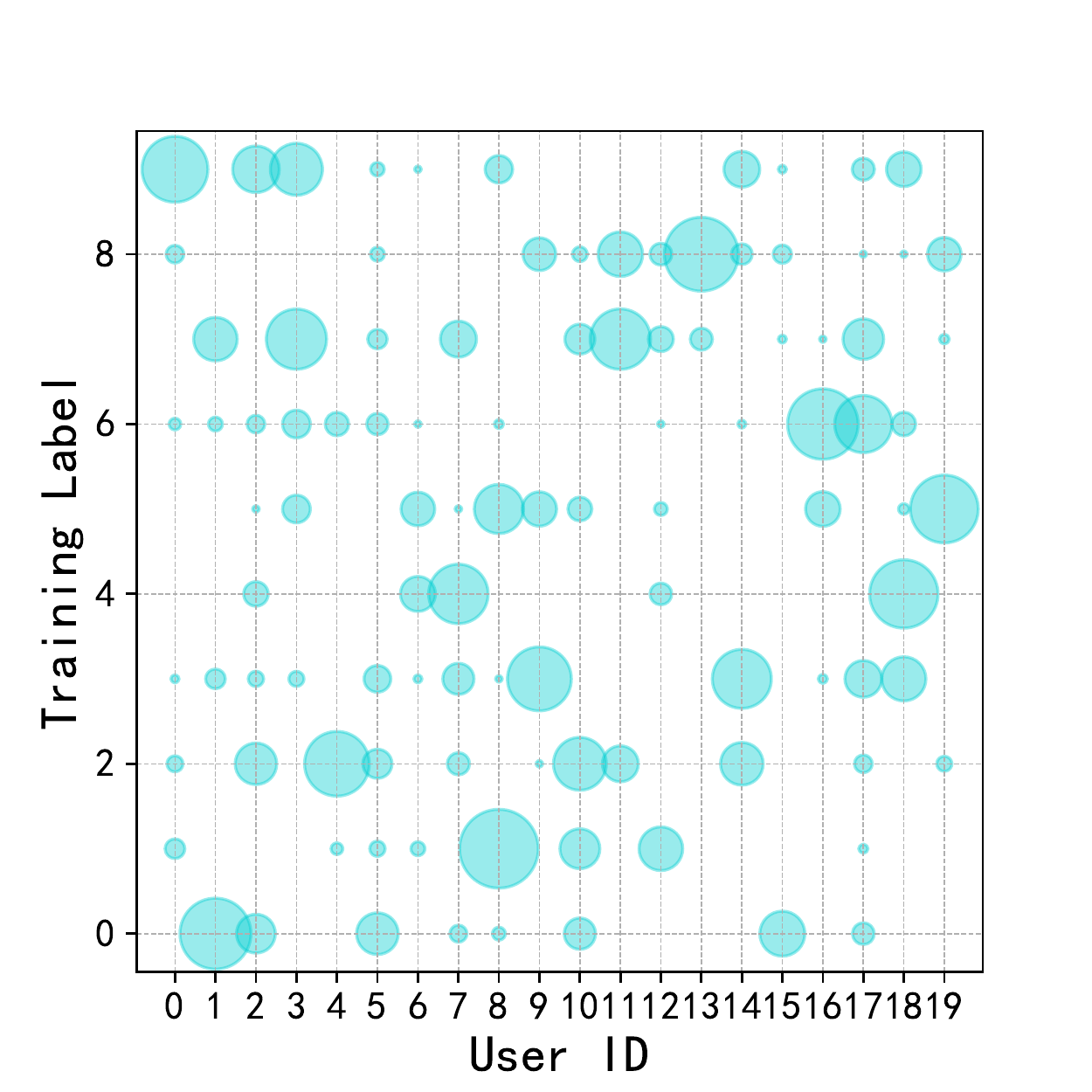}}
        \subcaption{\small MNIST$(\alpha=0.1)$}
        \end{subfigure}
        \hspace{0.1in}
        \begin{subfigure}[b]{0.23\textwidth}
            \centerline{\includegraphics[width=1.2\columnwidth]{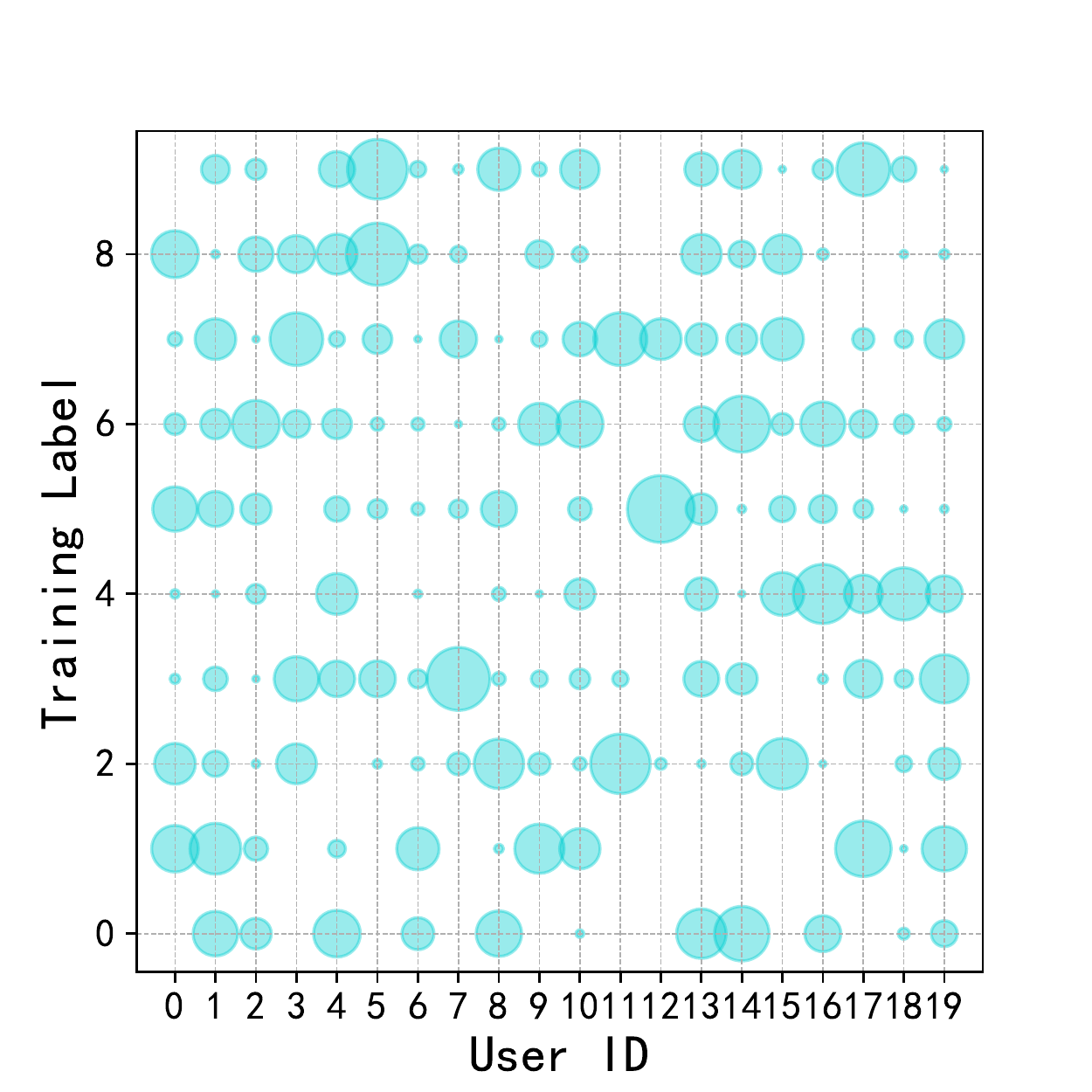}}
        \subcaption{\small MNIST$(\alpha=0.3)$}
        \end{subfigure}
         \hspace{0.1in}
        \begin{subfigure}[b]{0.23\textwidth}
            \centerline{\includegraphics[width=1.2\columnwidth]{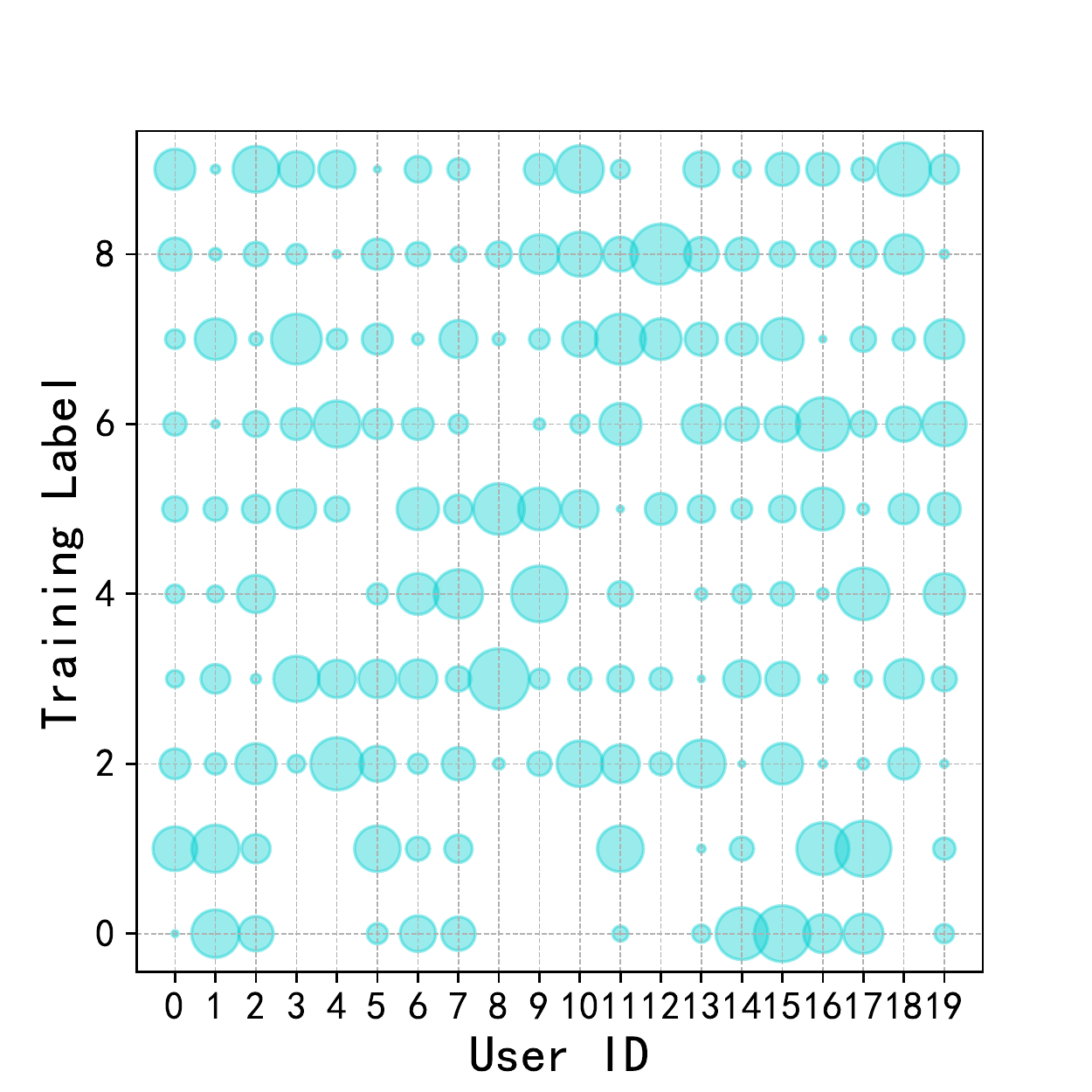}}
        \subcaption{\small MNIST$(\alpha=0.5)$}
        \end{subfigure}
          \hspace{0.1in}
        \begin{subfigure}[b]{0.23\textwidth}
            \centerline{\includegraphics[width=1.2\columnwidth]{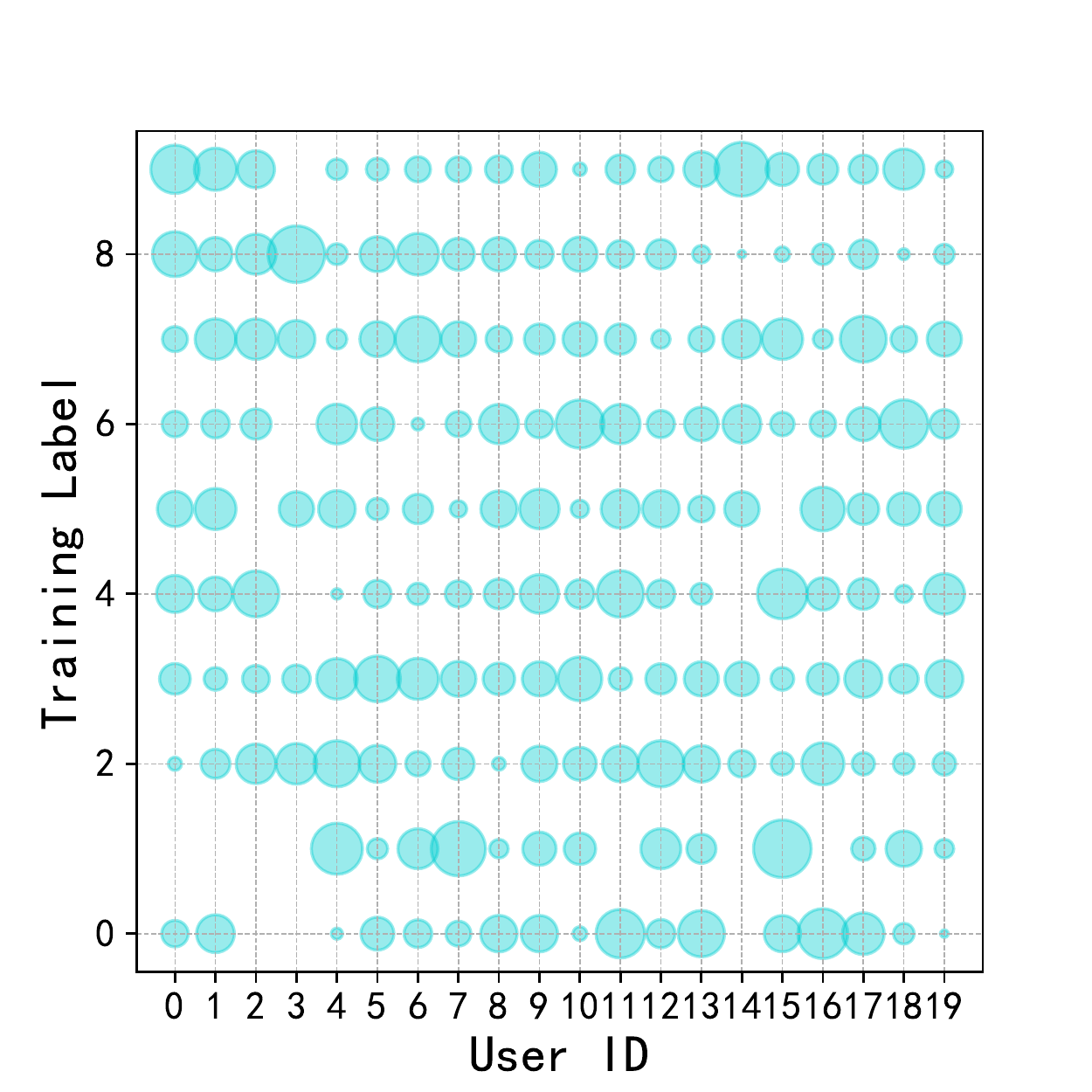}}
        \subcaption{\small MNIST$(\alpha=1.0)$}
        \end{subfigure}

        \begin{subfigure}[b]{0.23\textwidth}
            \centerline{\includegraphics[width=1.2\columnwidth]{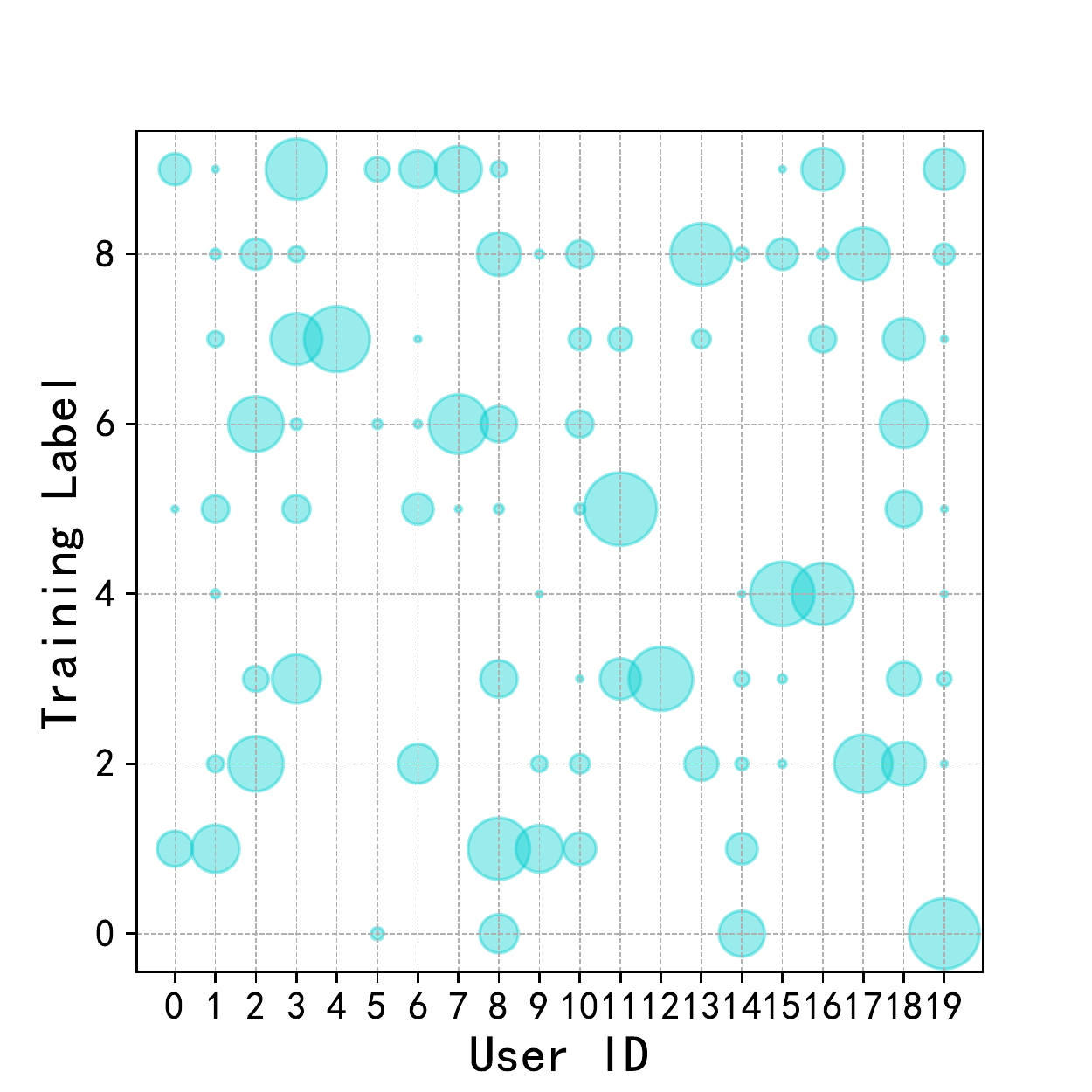}}
        \subcaption{\small CIFAR-10$(\alpha=0.1)$}
        \end{subfigure}
         \hspace{0.1in}
        \begin{subfigure}[b]{0.23\textwidth}
            \centerline{\includegraphics[width=1.2\columnwidth]{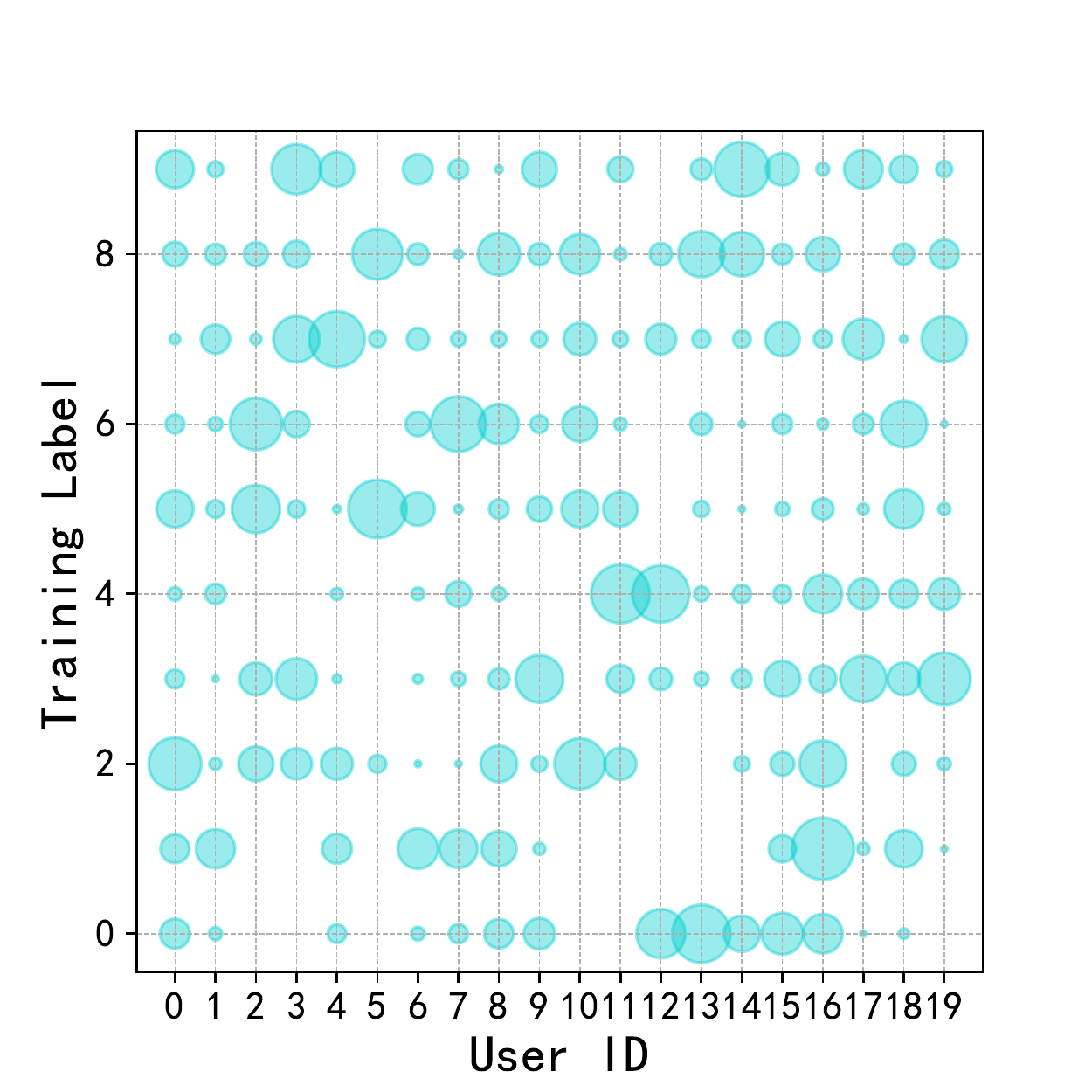}}
        \subcaption{\small CIFAR-10$(\alpha=0.3)$}
        \end{subfigure}
         \hspace{0.1in}
        \begin{subfigure}[b]{0.23\textwidth}
            \centerline{\includegraphics[width=1.2\columnwidth]{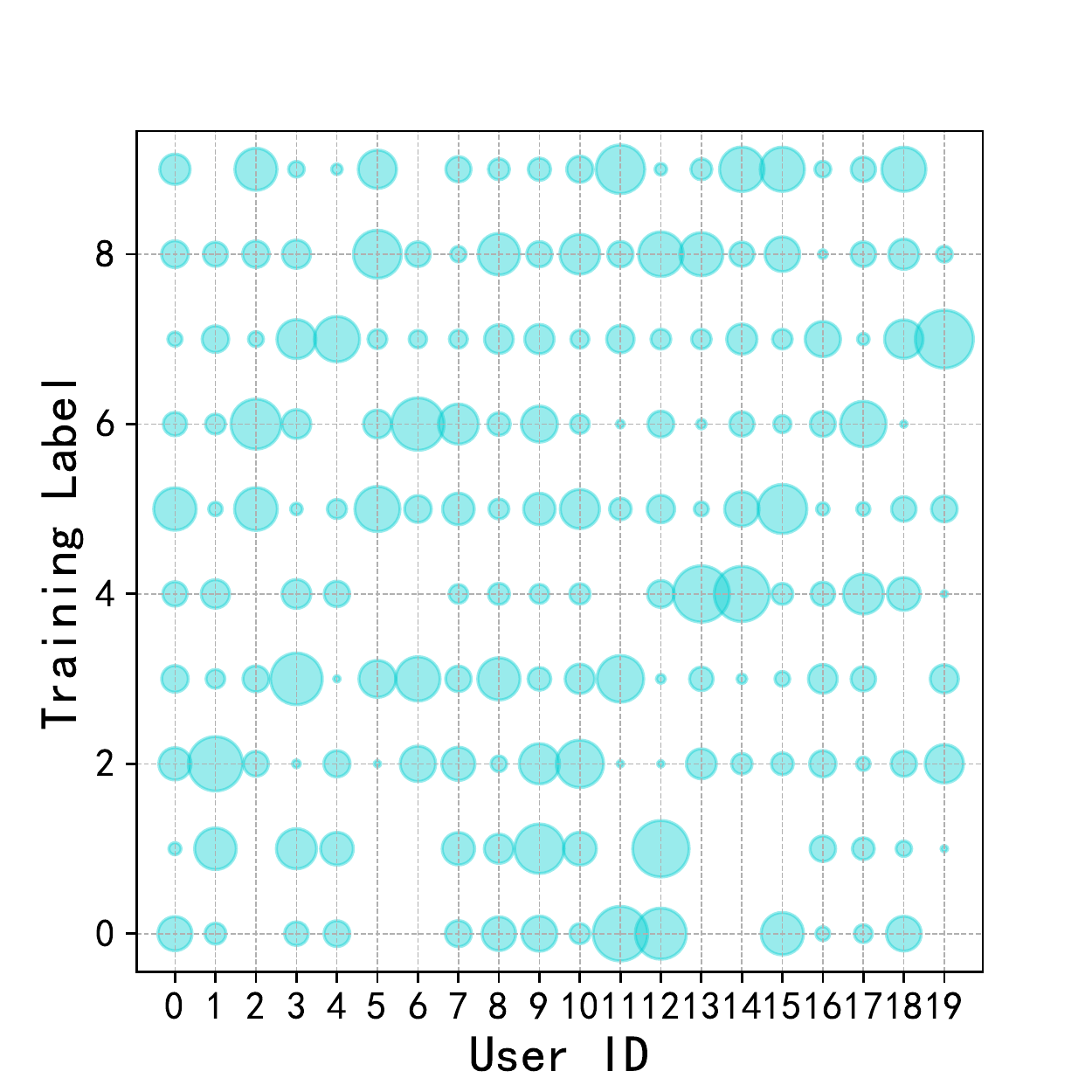}}
        \subcaption{\small CIFAR-10$(\alpha=0.5)$}
        \end{subfigure}
         \hspace{0.1in}
        \begin{subfigure}[b]{0.23\textwidth}
            \centerline{\includegraphics[width=1.2\columnwidth]{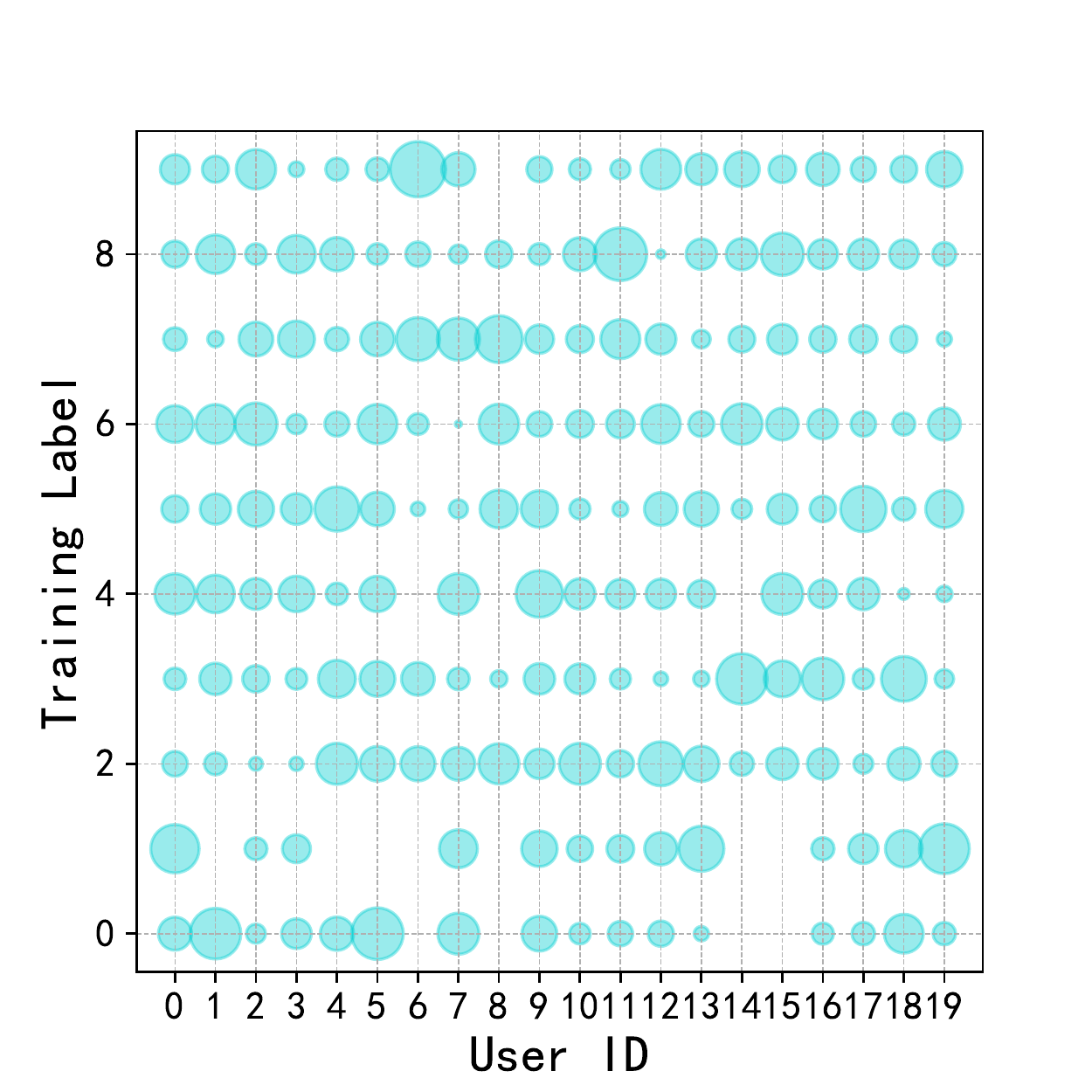}}
        \subcaption{\small CIFAR-10$(\alpha=1.0)$}
        \end{subfigure}

    \caption{Visualization of statistical heterogeneity among clients, where the $x$-axis indicates user IDs, the $y$-axis indicates class labels, and the size of scattered points indicates the number of training samples.}\label{fig:label-heterogeneity}
    \end{center}
\end{figure*}

\paragraph{Datasets.} We use MNIST~\cite{1998Gradient} and CIFAR-10~\cite{2009Learning} as benchmarks. To simulate the heterogeneous federated learning scenario, we follow the previous works~\cite{yurochkin2019bayesian,wang2020federated} that utilize 
Dirichlet distribution
$Dir(\alpha)$ to partition the training dataset and generate the corresponding test data for each client following the same distribution, in which a smaller $\alpha$ indicates the higher data heterogeneity. In our experiments, the number of total users is $20$. We also visualize the statistical heterogeneity of clients by adopting the visualization method in previous work~\cite{zhu2021data}, the results are shown in Figure \ref{fig:label-heterogeneity}.
\paragraph{Model.} For MNIST, we use the fully connected network which contains 3 FC layers, the FC layers are with 260, 200 hidden sizes, and 10 neurons for 10 classes as outputs, respectively. Since we adopt the binary training strategy, the activation function \textbf{sigmoid} used in binary classification is applied and maps outputs into the interval $[0,1]$. We normally apply softmax activation for the compared methods. For CIFAR-10, we use ConvNet~\cite{lecun1998gradient}, which contains 2 Conv layers and 3 FC layers. The Conv layers have 64 and 64 channels, respectively. The FC layers are with 120, 64 hidden sizes, and 10 neurons as outputs, respectively. Similar to the fully connected network used on MNIST, the activation function \textbf{sigmoid} is also applied.
\paragraph{Configuration.} Our method has four hyper-parameters: $m^{p}$,${m^{n}}$,$m^{nn}$ and $\sigma$. These parameters amongst different clients can be varied according to local imbalanced data. If we adopt a flexible parameter setting strategy for each client, the experimented results will be better, but we make them equal for simplicity. For CIFAR-10, we set them as $0.85$, $0.2$, $0.3$, and $2$, respectively. For MNIST, we set them as $0.75$, $0.25$, $0.3$, and $2$, respectively. Throughout
the experiments, we use the SGD optimizer with weight decay $1e-5$ and a $0.9$ momentum and the bath size is 64. For MNIST, the learning rate is $0.01$. For CIFAR-10, the learning rate is $0.1$. We train every method for $100$ rounds and $200$ rounds on MNIST and CIFAR-10, respectively. For the federated framework setting, the participation rate of clients is set as $0.5$, which means that random $10$ clients will be activated in each communication round. The local training epochs are set as $5$ for all the experiments.
\paragraph{Compared methods.} We follow the observation~\cite{chen2021bridging} that if we adopt personalized models in generic FL algorithms, they outperform most of the existing PFL algorithms. Hence, we select some challenging generic FL algorithms including FedAvg~\cite{mcmahan2017communication}, FedProx~\cite{li2020federated}, and Scaffold~\cite{karimireddy2020scaffold}. We evaluate their \textbf{personalized models} to obtain corresponding PFL accuracy of those generic FL algorithms. For PFL methods, LG\_FedAvg~\cite{liang2020think}, FedPer~\cite{2019Federated}, FedRep~\cite{collins2021exploiting} and FedRod~\cite{chen2021bridging} are selected as challenging PFL baselines. In particular, FedRod has double classifier layers to perform on the global model and personalized model, respectively. In our experiments, FedRod adopts its linear mode where only G-Head is aggregated at the server.

\begin{table*}[!t]

\caption{ PFL accuracy(\%) on MNIST and CIFAR-10 under different degrees of heterogeneity (0.1, 0.3, 0.5, 1.0). The underline highlight the best-performing compared approach.}
\centering
\label{tab1}
 \resizebox{\textwidth}{!}{
 \begin{tabular}{c|c|cccc|ccccc}
         \toprule
        {Dataset}& {Non-iid}&{Local only}&{FedAvg}&{FedProx}&{Scaffold}&{LG\_FedAvg}& {FedPer}&{FedRep}&{FedRod}&{FedABC}\\
        \midrule
        \multirow{4}{*}{MNIST}&{Dir(0.1)}&{97.1}&{99.1}&{98.4}&{99.3}&{99.2}&{99.1}&{98.9}&{
        \uline{99.3}}&\textbf{99.3}\\
         \multirow{4}{*}{}&{Dir(0.3)}    &{93.3}&{98.3}&{97.1}&{98.6}&{98.2}&{98.2}&{97.8}& \uline{98.7}&\textbf{98.7}\\

        \multirow{4}{*}{}&{Dir(0.5)}     &{92.1}&{98.1}&{96.9}&{98.5}&{98.1}&{98.0}&{97.6}& \uline{98.6}&{98.5}\\
        \multirow{4}{*}{}&{Dir(1.0)}     &{90.6}&{98.0}&{96.9}&{98.4}&{98.1}&{98.1}&{97.4}& \uline{98.5}&\textbf{98.7}\\

        \midrule
        \multirow{4}{*}{CIFAR-10}&{Dir(0.1)}&{86.0}&{91.0}&{91.1}& \uline{91.2}&{88.7}&{90.2}&{90.4}&{90.5}&{91.0}\\
         \multirow{4}{*}{}&{Dir(0.3)}      &{72.4}&{82.1}&{82.2}&{81.9}&{76.6}&{81.2}&{81.4}& \uline{82.2}&\textbf{83.3}\\

        \multirow{4}{*}{}&{Dir(0.5)}       &{67.8}&{79.8}&{79.3}&{79.5}&{72.8}&{78.8}&{79.1}& \uline{79.8}&\textbf{81.1}\\
        \multirow{4}{*}{}&{Dir(1.0)}       &{83.7}&{74.1}&{73.8}&{74.0}&{63.7}&{72.5}&{73.4}&\uline{74.1}&\textbf{76.1}\\
       \bottomrule


    \end{tabular}}
\end{table*}

\subsection{Evaluation criteria}
\paragraph{Evaluation of personalized model.}To simulate the Non-iid scenario in FL, each client has the local train set and the corresponding test set with the same Dirichlet distribution. For the evaluation of PFL, we use the accuracy of the local Non-iid test set and the formulation can be given by the following:
\begin{equation}
\small
\textbf{PFL-accuracy: } T_{\text{PFL}}=\frac{\sum_{i} {I(y_{j}=\hat{y_{j}};D_{\footnotesize \text{ Non\_iid\_test}}^{i};\boldsymbol{\theta}^{i})}}{\sum_{i}|D^{i}_{\footnotesize \text{ Non\_iid\_test}}|}
\label{eva-PFL}
\end{equation}
where $I(\cdot)$ is an indicator function ($I(E)=1$ if event $E$ is true, and $0$ otherwise). The test accuracy of PFL is obtained by the sum of all the local true predictions number divided by the sum of the number of all test sets. It is worth emphasizing that the local testing process utilizes the personalized model instead of the global model.
\paragraph{Evaluation of client drift.} We also propose an evaluation method to quantify the degree of client drift. Since the varied data distributions, the local training performs with biases towards the local classes and easily forgets other knowledge, including other classes' features or different features of the same class. Aiming at quantifying the degree of client drift, we can evaluate the personalized model on the local iid test set, which has all categories and each one has the same number of samples. This iid test set can guarantee the generalization and thus we can treat the resulted accuracy as the metric of client drift. In this evaluation, the low value indicates a high degree of drift while the great value indicates that local training does not completely forget other knowledge and has a low degree of client drift. The formulation can be given by the following:
\begin{equation}
\small
\textbf{PFL-accuracy: } T_{\text{PFL}}=\frac{\sum_{i} {I(y_{j}=\hat{y_{j}};D^{i}_{\footnotesize \textbf{iid\_test}};\boldsymbol{\theta}^{i})}}{\sum_{i}|D^{i}_{\footnotesize \textbf{iid\_test}}|}
\label{eva-PFL}
\end{equation}

\begin{table*}[!t]

\caption{Drift degree of different methods. Client drift is quantified by the accuracy of applying the personalized model on iid test set. A lower value indicates a larger degree of drift. The underline highlight the best-performing compared approach.}
\centering
\label{tab2}
 \resizebox{\textwidth}{!}{
 \begin{tabular}{c|c|cccc|ccccc}
         \toprule
        {Dataset}& {Non-iid}&{Local only}&{FedAvg}&{FedProx}&{Scaffold}&{LG\_FedAvg}& {FedPer}&{FedRep}&{FedRod}&{FedABC}\\
        \midrule
        \multirow{4}{*}{MNIST}&{Dir(0.1)}&{29.2}&{43.7}&{39.4}&\uline{76.8}&{43.9}&{43.7}&{42.9}&{{66.9}}&\textbf{76.5}\\
         \multirow{4}{*}{}&{Dir(0.3)}  &{44.2}&{68.2}&{62.2}&\uline{90.2}&{68.1}&{68.2}&{66.3}&{85.4}&\textbf{91.1}\\

        \multirow{4}{*}{}&{Dir(0.5)}     &{58.0}&{78.7}&{74.0}&\uline{94.0}&{78.7}&{78.8}&{77.1}&{90.8}&\textbf{95.1}\\
        \multirow{4}{*}{}&{Dir(1.0)}     &{69.5}&{87.6}&{84.9}&\uline{97.0}&{87.6}&{87.6}&{86.4}&{95.3}&\textbf{97.2}\\

        \midrule
        \multirow{4}{*}{CIFAR-10}&{Dir(0.1)}&{16.7}&{37.0}&{37.8}&\uline{39.6}&{20.6}&{38.6}&{23.5}&{31.5}&\textbf{34.5}\\
         \multirow{4}{*}{}&{Dir(0.3)}      &{23.3}&{56.4}&{56.3}&\uline{56.1}&{32.1}&{42.0}&{41.6}&{51.4}&\textbf{54.7}\\

        \multirow{4}{*}{}&{Dir(0.5)}       &{25.6}&{60.0}&{59.9}&\uline{60.0}&{35.4}&{47.0}&{46.5}&{55.5}&\textbf{59.0}\\
        \multirow{4}{*}{}&{Dir(1.0)}       &{14.4}&{64.8}&{64.2}&\uline{65.2}&{39.7}&{52.7}&{53.8}&{61.5}&\textbf{64.0}\\
       \bottomrule
    \end{tabular}}
\label{client_drift}
\end{table*}

\subsection{Results and analysis}
We present the results in Tables \ref{tab1} \&\ref{tab2}, underline and bold fonts highlight the best baseline/our approach. From the experimented results, we observe that: 1) the advanced local training of generic FL can achieve promising PFL accuracy and even outperforms other existing PFL algorithms, especially in the huge heterogeneity setting(i.e.,0.1). The reason behind this is that the data distribution of test data is the same as the training data. To understand it better, we can assume that each client only accesses a single class, the PFL accuracy in this situation will close to $100\%$; 2) the accuracy of PFL decrease with the level of heterogeneity decrease because the local training is hard to cater for all classes inside clients; 3) FedRod achieves the best performance in baselines. FedRod has double classifiers for global paradigm and personalized paradigm and thus is more robust. 4) Regard to the PFL accuracy, our FedABC achieves promising performance and outperforms other baselines. 4) Regard to the client drift, Scaffold uses control variates to correct for the `client-drift' in its local updates and thus achieves the best performance. Our FedABC also achieves nice performance compared with other methods, especially in PFL baselines.
\paragraph{Ablation study}
\begin{figure}[!t]
    \begin{center}
    \includegraphics[width=0.8\columnwidth]{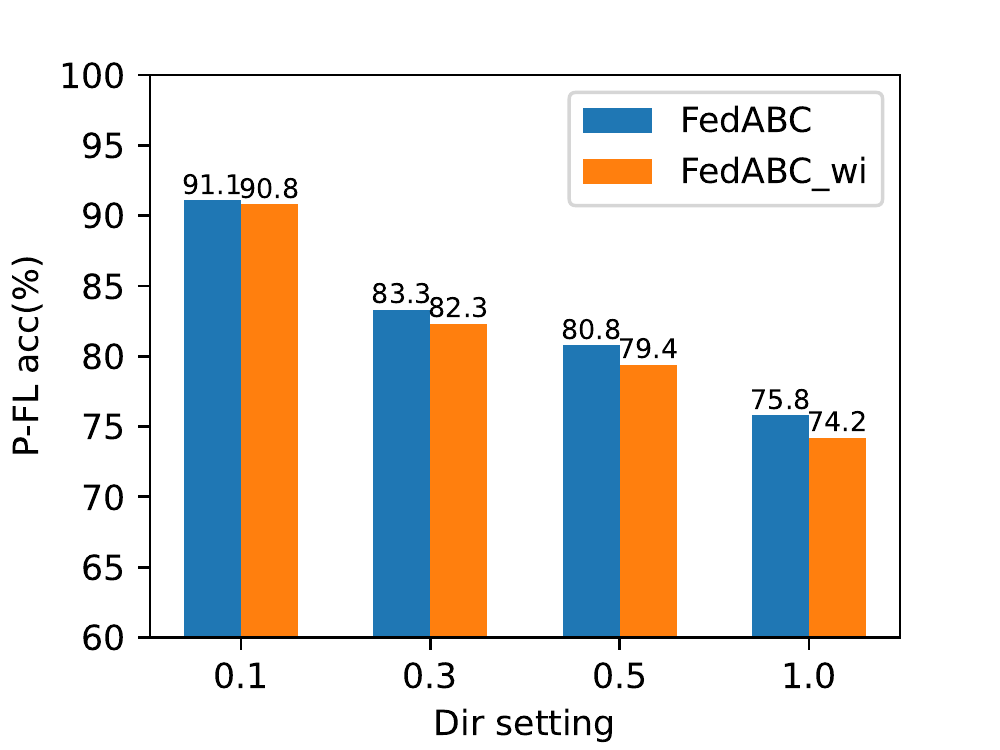}
    \end{center}
    \caption{An ablation study of the proposed FedABC method, where the orange bars (FedABC$\_$wi) are the performance of our method without adopting the designed binary classification loss that incorporates the under-sampling and hard sample mining strategies.}
    \label{Ablation study}
\end{figure}
We study the necessity of the component for solving the imbalanced problem in our binary classification. We conduct new experiments on CIFAR-10. We train $200$ epochs for the compared method FedABC\_wi without adopting our imbalanced training technique(under-sampling and hard sample mining), the experimented results are shown in Figure \ref{Ablation study}. From the figure, we can find that: 1) Adopting our imbalanced training technique can effectively improve PFL performance. The specific accuracy improvement is $0.3\%$, $1.0\%$, $1.4\%$, $1.6\%$ for the heterogeneity setting $0.1$, $0.3$, $0.5$, $1.0$, respectively. The positive samples and negative samples sometimes are seriously imbalanced in our binary classification and the efficient imbalanced training technique can alleviate this problem and thus generate improvement; 2) The improvement decreases with the degree of heterogeneity increases. In the huge heterogeneity setting(e.g. Dir(0.1)), the improvement is slight but significant for other settings(e.g. Dir(1.0)). The reason behind it is that re-balancing local data distribution may generate a trade-off problem between the local personality and the imbalanced problem. Since the data distribution of the test set is the same as the train set, excessive re-balancing may inevitably hurt the local personality. Meanwhile, training with a serious imbalanced problem can also impact the classifier decision on minority classes, especially in our binary training strategy.

\section{Conclusion}
In this paper, we investigate some extraordinary Non-IID situations in federated learning, where the data distributions among clients are imbalanced and some classes even have no positive samples. These issues are alleviated by constructing a binary classification problem for each category instead of adopting the popular Softmax function. This training strategy may aggravate the class imbalance problem and thus a novel loss function that incorporates the under-sampling and hard sample mining are further designed. Extensive experiments are conducted on two popular datasets, and the results show that our FedABC can significantly improve the PFL performance in diverse heterogeneity settings. The limitation of our FedABC may be the possible conflictions given hundreds and thousands of categories. In the future, we intend to integrate with the communication strategies to further improve the performance and conduct experiments on more large-scale datasets.

\section{Acknowledgments}
This work was supported by the National Key Research and Development Program of China under Grant 2021YFC3300200, the Special Fund of Hubei Luojia Laboratory under Grant 220100014, the National Natural Science Foundation of China (Grant No. 62276195 and 62272354), and the Nation Research Foundation, Prime Minister's Office, Singapore under its Energy Programme (EP Award No. NRF2017EWT-EP003-023) administrated by the Energy Market Authority of Singapore.

\normalem
\bibliography{ref}

\begin{thebibliography}{56}
\providecommand{\natexlab}[1]{#1}

\bibitem[{Acar et~al.(2021)Acar, Zhao, Navarro, Mattina, Whatmough, and
  Saligrama}]{acar2021federated}
Acar, D. A.~E.; Zhao, Y.; Navarro, R.~M.; Mattina, M.; Whatmough, P.~N.; and
  Saligrama, V. 2021.
\newblock Federated learning based on dynamic regularization.
\newblock \emph{arXiv preprint arXiv:2111.04263}.

\bibitem[{Arivazhagan et~al.(2019)Arivazhagan, Aggarwal, Singh, and
  Choudhary}]{2019Federated}
Arivazhagan, M.~G.; Aggarwal, V.; Singh, A.~K.; and Choudhary, S. 2019.
\newblock Federated Learning with Personalization Layers.

\bibitem[{Aurelio et~al.(2019)Aurelio, de~Almeida, de~Castro, and
  Braga}]{aurelio2019learning}
Aurelio, Y.~S.; de~Almeida, G.~M.; de~Castro, C.~L.; and Braga, A.~P. 2019.
\newblock Learning from imbalanced data sets with weighted cross-entropy
  function.
\newblock \emph{Neural processing letters}, 50(2): 1937--1949.

\bibitem[{Buda, Maki, and Mazurowski(2018)}]{buda2018systematic}
Buda, M.; Maki, A.; and Mazurowski, M.~A. 2018.
\newblock A systematic study of the class imbalance problem in convolutional
  neural networks.
\newblock \emph{Neural networks}, 106: 249--259.

\bibitem[{Chen et~al.(2018)Chen, Luo, Dong, Li, and He}]{chen2018federated}
Chen, F.; Luo, M.; Dong, Z.; Li, Z.; and He, X. 2018.
\newblock Federated meta-learning with fast convergence and efficient
  communication.
\newblock \emph{arXiv preprint arXiv:1802.07876}.

\bibitem[{Chen and Chao(2021)}]{chen2021bridging}
Chen, H.-Y.; and Chao, W.-L. 2021.
\newblock On bridging generic and personalized federated learning for image
  classification.
\newblock In \emph{International Conference on Learning Representations}.

\bibitem[{Collins et~al.(2021)Collins, Hassani, Mokhtari, and
  Shakkottai}]{collins2021exploiting}
Collins, L.; Hassani, H.; Mokhtari, A.; and Shakkottai, S. 2021.
\newblock Exploiting shared representations for personalized federated
  learning.
\newblock In \emph{International Conference on Machine Learning}, 2089--2099.
  PMLR.

\bibitem[{Dai et~al.(2022)Dai, Shen, He, Tian, and Tao}]{dai2022dispfl}
Dai, R.; Shen, L.; He, F.; Tian, X.; and Tao, D. 2022.
\newblock DisPFL: Towards Communication-Efficient Personalized Federated
  Learning via Decentralized Sparse Training.
\newblock \emph{arXiv preprint arXiv:2206.00187}.

\bibitem[{Drummond, Holte et~al.(2003)}]{drummond2003c4}
Drummond, C.; Holte, R.~C.; et~al. 2003.
\newblock C4. 5, class imbalance, and cost sensitivity: why under-sampling
  beats over-sampling.
\newblock In \emph{Workshop on learning from imbalanced datasets II},
  volume~11, 1--8. Citeseer.

\bibitem[{Geifman and El-Yaniv(2017)}]{geifman2017deep}
Geifman, Y.; and El-Yaniv, R. 2017.
\newblock Deep active learning over the long tail.
\newblock \emph{arXiv preprint arXiv:1711.00941}.

\bibitem[{Guo et~al.(2017)Guo, Pleiss, Sun, and
  Weinberger}]{guo2017calibration}
Guo, C.; Pleiss, G.; Sun, Y.; and Weinberger, K.~Q. 2017.
\newblock On calibration of modern neural networks.
\newblock In \emph{International conference on machine learning}, 1321--1330.
  PMLR.

\bibitem[{Hanzely et~al.(2020)Hanzely, Hanzely, Horv{\'a}th, and
  Richt{\'a}rik}]{hanzely2020lower}
Hanzely, F.; Hanzely, S.; Horv{\'a}th, S.; and Richt{\'a}rik, P. 2020.
\newblock Lower bounds and optimal algorithms for personalized federated
  learning.
\newblock \emph{Advances in Neural Information Processing Systems}, 33:
  2304--2315.

\bibitem[{Hanzely, Zhao, and Kolar(2021)}]{hanzely2021personalized}
Hanzely, F.; Zhao, B.; and Kolar, M. 2021.
\newblock Personalized federated learning: A unified framework and universal
  optimization techniques.
\newblock \emph{arXiv preprint arXiv:2102.09743}.

\bibitem[{He and Garcia(2009)}]{he2009learning}
He, H.; and Garcia, E.~A. 2009.
\newblock Learning from imbalanced data.
\newblock \emph{IEEE Transactions on knowledge and data engineering}, 21(9):
  1263--1284.

\bibitem[{Hong et~al.(2021)Hong, Han, Choi, Seo, Kim, and
  Chang}]{hong2021disentangling}
Hong, Y.; Han, S.; Choi, K.; Seo, S.; Kim, B.; and Chang, B. 2021.
\newblock Disentangling label distribution for long-tailed visual recognition.
\newblock In \emph{Proceedings of the IEEE/CVF conference on computer vision
  and pattern recognition}, 6626--6636.

\bibitem[{Hosseini et~al.(2021)Hosseini, Park, Yun, Louizos, Soriaga, and
  Welling}]{hosseini2021federated}
Hosseini, H.; Park, H.; Yun, S.; Louizos, C.; Soriaga, J.; and Welling, M.
  2021.
\newblock Federated Learning of User Verification Models Without Sharing
  Embeddings.
\newblock \emph{arXiv preprint arXiv:2104.08776}.

\bibitem[{Hsieh et~al.(2020)Hsieh, Phanishayee, Mutlu, and
  Gibbons}]{hsieh2020non}
Hsieh, K.; Phanishayee, A.; Mutlu, O.; and Gibbons, P. 2020.
\newblock The non-iid data quagmire of decentralized machine learning.
\newblock In \emph{International Conference on Machine Learning}, 4387--4398.
  PMLR.

\bibitem[{Huang et~al.(2016)Huang, Li, Loy, and Tang}]{huang2016learning}
Huang, C.; Li, Y.; Loy, C.~C.; and Tang, X. 2016.
\newblock Learning deep representation for imbalanced classification.
\newblock In \emph{Proceedings of the IEEE conference on computer vision and
  pattern recognition}, 5375--5384.

\bibitem[{Huang et~al.(2022{\natexlab{a}})Huang, Lin, Shen, Li, and
  Zomaya}]{huang2022stochastic}
Huang, T.; Lin, W.; Shen, L.; Li, K.; and Zomaya, A.~Y. 2022{\natexlab{a}}.
\newblock Stochastic client selection for federated learning with volatile
  clients.
\newblock \emph{IEEE Internet of Things Journal}.

\bibitem[{Huang et~al.(2022{\natexlab{b}})Huang, Liu, Shen, He, Lin, and
  Tao}]{huang2022achieving}
Huang, T.; Liu, S.; Shen, L.; He, F.; Lin, W.; and Tao, D. 2022{\natexlab{b}}.
\newblock Achieving Personalized Federated Learning with Sparse Local Models.
\newblock \emph{arXiv preprint arXiv:2201.11380}.

\bibitem[{Jang, Gu, and Poole(2016)}]{jang2016categorical}
Jang, E.; Gu, S.; and Poole, B. 2016.
\newblock Categorical reparameterization with gumbel-softmax.
\newblock \emph{arXiv preprint arXiv:1611.01144}.

\bibitem[{Japkowicz and Stephen(2002)}]{japkowicz2002class}
Japkowicz, N.; and Stephen, S. 2002.
\newblock The class imbalance problem: A systematic study.
\newblock \emph{Intelligent data analysis}, 6(5): 429--449.

\bibitem[{Kang et~al.(2019)Kang, Xie, Rohrbach, Yan, Gordo, Feng, and
  Kalantidis}]{kang2019decoupling}
Kang, B.; Xie, S.; Rohrbach, M.; Yan, Z.; Gordo, A.; Feng, J.; and Kalantidis,
  Y. 2019.
\newblock Decoupling representation and classifier for long-tailed recognition.
\newblock \emph{arXiv preprint arXiv:1910.09217}.

\bibitem[{Karimireddy et~al.(2020)Karimireddy, Kale, Mohri, Reddi, Stich, and
  Suresh}]{karimireddy2020scaffold}
Karimireddy, S.~P.; Kale, S.; Mohri, M.; Reddi, S.; Stich, S.; and Suresh,
  A.~T. 2020.
\newblock Scaffold: Stochastic controlled averaging for federated learning.
\newblock In \emph{International Conference on Machine Learning}, 5132--5143.
  PMLR.

\bibitem[{Katharopoulos and Fleuret(2018)}]{katharopoulos2018not}
Katharopoulos, A.; and Fleuret, F. 2018.
\newblock Not all samples are created equal: Deep learning with importance
  sampling.
\newblock In \emph{International conference on machine learning}, 2525--2534.
  PMLR.

\bibitem[{Khodak, Balcan, and Talwalkar(2019)}]{khodak2019adaptive}
Khodak, M.; Balcan, M.-F.~F.; and Talwalkar, A.~S. 2019.
\newblock Adaptive gradient-based meta-learning methods.
\newblock \emph{Advances in Neural Information Processing Systems}, 32.

\bibitem[{Krizhevsky and Hinton(2009)}]{2009Learning}
Krizhevsky, A.; and Hinton, G. 2009.
\newblock Learning multiple layers of features from tiny images.
\newblock \emph{Handbook of Systemic Autoimmune Diseases}, 1(4).

\bibitem[{Lecun and Bottou(1998)}]{1998Gradient}
Lecun, Y.; and Bottou, L. 1998.
\newblock Gradient-based learning applied to document recognition.
\newblock \emph{Proceedings of the IEEE}, 86(11): 2278--2324.

\bibitem[{LeCun et~al.(1998)LeCun, Bottou, Bengio, and
  Haffner}]{lecun1998gradient}
LeCun, Y.; Bottou, L.; Bengio, Y.; and Haffner, P. 1998.
\newblock Gradient-based learning applied to document recognition.
\newblock \emph{Proceedings of the IEEE}, 86(11): 2278--2324.

\bibitem[{Li et~al.(2021{\natexlab{a}})Li, Diao, Chen, and
  He}]{li2021federated}
Li, Q.; Diao, Y.; Chen, Q.; and He, B. 2021{\natexlab{a}}.
\newblock Federated learning on non-iid data silos: An experimental study.
\newblock \emph{arXiv preprint arXiv:2102.02079}.

\bibitem[{Li et~al.(2021{\natexlab{b}})Li, Hu, Beirami, and
  Smith}]{li2021ditto}
Li, T.; Hu, S.; Beirami, A.; and Smith, V. 2021{\natexlab{b}}.
\newblock Ditto: Fair and robust federated learning through personalization.
\newblock In \emph{International Conference on Machine Learning}, 6357--6368.
  PMLR.

\bibitem[{Li et~al.(2020)Li, Sahu, Talwalkar, and Smith}]{li2020federated}
Li, T.; Sahu, A.~K.; Talwalkar, A.; and Smith, V. 2020.
\newblock Federated learning: Challenges, methods, and future directions.
\newblock \emph{IEEE Signal Processing Magazine}, 37(3): 50--60.

\bibitem[{Liang et~al.(2020)Liang, Liu, Ziyin, Allen, Auerbach, Brent,
  Salakhutdinov, and Morency}]{liang2020think}
Liang, P.~P.; Liu, T.; Ziyin, L.; Allen, N.~B.; Auerbach, R.~P.; Brent, D.;
  Salakhutdinov, R.; and Morency, L.-P. 2020.
\newblock Think locally, act globally: Federated learning with local and global
  representations.
\newblock \emph{arXiv preprint arXiv:2001.01523}.

\bibitem[{Lin et~al.(2017)Lin, Goyal, Girshick, He, and
  Doll{\'a}r}]{lin2017focal}
Lin, T.-Y.; Goyal, P.; Girshick, R.; He, K.; and Doll{\'a}r, P. 2017.
\newblock Focal loss for dense object detection.
\newblock In \emph{Proceedings of the IEEE international conference on computer
  vision}, 2980--2988.

\bibitem[{Liu et~al.(2022)Liu, Lou, Wang, Xi, Shen, and Yan}]{liu2022deep}
Liu, C.; Lou, C.; Wang, R.; Xi, A.~Y.; Shen, L.; and Yan, J. 2022.
\newblock Deep neural network fusion via graph matching with applications to
  model ensemble and federated learning.
\newblock In \emph{International Conference on Machine Learning}, 13857--13869.
  PMLR.

\bibitem[{McMahan et~al.(2017)McMahan, Moore, Ramage, Hampson, and
  y~Arcas}]{mcmahan2017communication}
McMahan, B.; Moore, E.; Ramage, D.; Hampson, S.; and y~Arcas, B.~A. 2017.
\newblock Communication-efficient learning of deep networks from decentralized
  data.
\newblock In \emph{Artificial intelligence and statistics}, 1273--1282. PMLR.

\bibitem[{Ren et~al.(2020)Ren, Yu, Ma, Zhao, Yi et~al.}]{ren2020balanced}
Ren, J.; Yu, C.; Ma, X.; Zhao, H.; Yi, S.; et~al. 2020.
\newblock Balanced meta-softmax for long-tailed visual recognition.
\newblock \emph{Advances in neural information processing systems}, 33:
  4175--4186.

\bibitem[{Rifkin and Klautau(2004)}]{rifkin2004defense}
Rifkin, R.; and Klautau, A. 2004.
\newblock In defense of one-vs-all classification.
\newblock \emph{The Journal of Machine Learning Research}, 5: 101--141.

\bibitem[{Schroff, Kalenichenko, and Philbin(2015)}]{schroff2015facenet}
Schroff, F.; Kalenichenko, D.; and Philbin, J. 2015.
\newblock Facenet: A unified embedding for face recognition and clustering.
\newblock In \emph{Proceedings of the IEEE conference on computer vision and
  pattern recognition}, 815--823.

\bibitem[{Shen, Lin, and Huang(2016)}]{shen2016relay}
Shen, L.; Lin, Z.; and Huang, Q. 2016.
\newblock Relay backpropagation for effective learning of deep convolutional
  neural networks.
\newblock In \emph{European conference on computer vision}, 467--482. Springer.

\bibitem[{Smith et~al.(2017)Smith, Chiang, Sanjabi, and
  Talwalkar}]{smith2017federated}
Smith, V.; Chiang, C.-K.; Sanjabi, M.; and Talwalkar, A.~S. 2017.
\newblock Federated multi-task learning.
\newblock \emph{Advances in neural information processing systems}, 30.

\bibitem[{Sun et~al.(2021)Sun, Huo, Yang, and Bai}]{sun2021partialfed}
Sun, B.; Huo, H.; Yang, Y.; and Bai, B. 2021.
\newblock Partialfed: Cross-domain personalized federated learning via partial
  initialization.
\newblock \emph{Advances in Neural Information Processing Systems}, 34:
  23309--23320.

\bibitem[{T~Dinh, Tran, and Nguyen(2020)}]{t2020personalized}
T~Dinh, C.; Tran, N.; and Nguyen, J. 2020.
\newblock Personalized federated learning with moreau envelopes.
\newblock \emph{Advances in Neural Information Processing Systems}, 33:
  21394--21405.

\bibitem[{Van~Horn and Perona(2017)}]{van2017devil}
Van~Horn, G.; and Perona, P. 2017.
\newblock The devil is in the tails: Fine-grained classification in the wild.
\newblock \emph{arXiv preprint arXiv:1709.01450}.

\bibitem[{Wang et~al.(2020)Wang, Yurochkin, Sun, Papailiopoulos, and
  Khazaeni}]{wang2020federated}
Wang, H.; Yurochkin, M.; Sun, Y.; Papailiopoulos, D.; and Khazaeni, Y. 2020.
\newblock Federated learning with matched averaging.
\newblock \emph{arXiv preprint arXiv:2002.06440}.

\bibitem[{Wang et~al.(2019)Wang, Mathews, Kiddon, Eichner, Beaufays, and
  Ramage}]{wang2019federated}
Wang, K.; Mathews, R.; Kiddon, C.; Eichner, H.; Beaufays, F.; and Ramage, D.
  2019.
\newblock Federated evaluation of on-device personalization.
\newblock \emph{arXiv preprint arXiv:1910.10252}.

\bibitem[{Wen et~al.(2021)Wen, Liu, Weller, Raj, and
  Singh}]{wen2021sphereface2}
Wen, Y.; Liu, W.; Weller, A.; Raj, B.; and Singh, R. 2021.
\newblock Sphereface2: Binary classification is all you need for deep face
  recognition.
\newblock \emph{arXiv preprint arXiv:2108.01513}.

\bibitem[{Wu et~al.(2017)Wu, Manmatha, Smola, and Krahenbuhl}]{wu2017sampling}
Wu, C.-Y.; Manmatha, R.; Smola, A.~J.; and Krahenbuhl, P. 2017.
\newblock Sampling matters in deep embedding learning.
\newblock In \emph{Proceedings of the IEEE international conference on computer
  vision}, 2840--2848.

\bibitem[{Xu et~al.(2021)Xu, Glicksberg, Su, Walker, Bian, and
  Wang}]{xu2021federated}
Xu, J.; Glicksberg, B.~S.; Su, C.; Walker, P.; Bian, J.; and Wang, F. 2021.
\newblock Federated learning for healthcare informatics.
\newblock \emph{Journal of Healthcare Informatics Research}, 5(1): 1--19.

\bibitem[{Yen and Lee(2009)}]{yen2009cluster}
Yen, S.-J.; and Lee, Y.-S. 2009.
\newblock Cluster-based under-sampling approaches for imbalanced data
  distributions.
\newblock \emph{Expert Systems with Applications}, 36(3): 5718--5727.

\bibitem[{Yurochkin et~al.(2019)Yurochkin, Agarwal, Ghosh, Greenewald, Hoang,
  and Khazaeni}]{yurochkin2019bayesian}
Yurochkin, M.; Agarwal, M.; Ghosh, S.; Greenewald, K.; Hoang, N.; and Khazaeni,
  Y. 2019.
\newblock Bayesian nonparametric federated learning of neural networks.
\newblock In \emph{International Conference on Machine Learning}, 7252--7261.
  PMLR.

\bibitem[{Zhang et~al.(2022)Zhang, Shen, Ding, Tao, and Duan}]{zhang2022fine}
Zhang, L.; Shen, L.; Ding, L.; Tao, D.; and Duan, L.-Y. 2022.
\newblock Fine-tuning global model via data-free knowledge distillation for
  non-iid federated learning.
\newblock In \emph{Proceedings of the IEEE/CVF Conference on Computer Vision
  and Pattern Recognition}, 10174--10183.

\bibitem[{Zhao et~al.(2018)Zhao, Li, Lai, Suda, Civin, and
  Chandra}]{zhao2018federated}
Zhao, Y.; Li, M.; Lai, L.; Suda, N.; Civin, D.; and Chandra, V. 2018.
\newblock Federated learning with non-iid data.
\newblock \emph{arXiv preprint arXiv:1806.00582}.

\bibitem[{Zheng et~al.(2022)Zheng, Zhou, Sun, Wang, Liu, and
  Li}]{zheng2022applications}
Zheng, Z.; Zhou, Y.; Sun, Y.; Wang, Z.; Liu, B.; and Li, K. 2022.
\newblock Applications of federated learning in smart cities: recent advances,
  taxonomy, and open challenges.
\newblock \emph{Connection Science}, 34(1): 1--28.

\bibitem[{Zhu, Hong, and Zhou(2021)}]{zhu2021data}
Zhu, Z.; Hong, J.; and Zhou, J. 2021.
\newblock Data-free knowledge distillation for heterogeneous federated
  learning.
\newblock In \emph{International Conference on Machine Learning}, 12878--12889.
  PMLR.

\bibitem[{Zou et~al.(2018)Zou, Yu, Kumar, and Wang}]{zou2018unsupervised}
Zou, Y.; Yu, Z.; Kumar, B.; and Wang, J. 2018.
\newblock Unsupervised domain adaptation for semantic segmentation via
  class-balanced self-training.
\newblock In \emph{Proceedings of the European conference on computer vision
  (ECCV)}, 289--305.

\end{thebibliography}

\end{document}